\documentclass[preprint,12pt,authoryear]{elsarticle}

\usepackage{amssymb}
\usepackage{booktabs} 
\usepackage{multirow} 
\usepackage{geometry} 
\usepackage{tabularx} 
\usepackage{subcaption} 
\usepackage{caption}

\usepackage{amsmath}
\usepackage{hyperref}
\usepackage{adjustbox}

\journal{Nuclear Physics B}

\begin{document}

\begin{frontmatter}



\title{MDSAM:Memory-Driven Sparse Attention Matrix for LVLMs Hallucination Mitigation} 


\author{Shuaiye Lu, Linjiang Zhou, Xiaochuan Shi} 

\affiliation{organization={Wuhan University},
            addressline={shuaiyelu@whu.edu.cn}, 
            }

\begin{abstract}
Hallucinations in large vision-language models (LVLMs) often stem from the model's sensitivity to image tokens during decoding, as evidenced by attention peaks observed when generating both real and hallucinated entities. To address this, we propose Memory-Driven Sparse Attention Matrix (MDSAM) , a novel training-free approach that dynamically captures and refines the attention allocated to image tokens at each layer. MDSAM memorizes attention patterns and activates updates through alignment during decoding, enhancing focus on relevant image tokens while effectively reducing hallucinations. We evaluate MDSAM on multiple benchmarks for tasks such as image captioning and visual question answering, demonstrating its ability to consistently reduce hallucinations and improve reliability. Compatible with various LVLM architectures, MDSAM highlights its adaptability and effectiveness in mitigating hallucinations without requiring additional training or external tools.
\end{abstract}






\end{frontmatter}



\section{Introduction}
Large Vision-Language Models (LVLMs) (\cite{liu2023visual,liu2024improved,zhu2023minigpt,ye2023mplug,gong2023multimodal,bai2023qwenvl,li2023blip,lu2024deepseek,wu2024deepseek})
have recently shown remarkable progress in integrating visual and textual information, enabling breakthroughs in tasks like visual question answering(\cite{chen2023frugalgpt,team2023gemini,hurst2024gpt}) 
and image captioning(\cite{li2019visualbert,wang2023caption}). 
These models leverage pre-trained visual encoders and transformer-based language decoders to bridge modalities, achieving impressive versatility(\cite{radford2021learning}).

Despite their advancements, LVLMs still grapple with the critical issue of hallucination(\cite{liu2023mitigating,leng2024mitigating,deng2024seeing}, where generated text diverges from the actual visual input (\cite{liu2024improved,wang2024qwen2}). This challenge hinders their reliability in real-world applications. Similarly, LVLMs aim to extend language models for multimodal tasks but face comparable concerns regarding accuracy and consistency (\cite{liu2024survey}). Addressing these limitations is essential for advancing their practical utility and ensuring robust performance across diverse scenarios.

\begin{figure}
  \centering
  \includegraphics[width=0.9\textwidth]{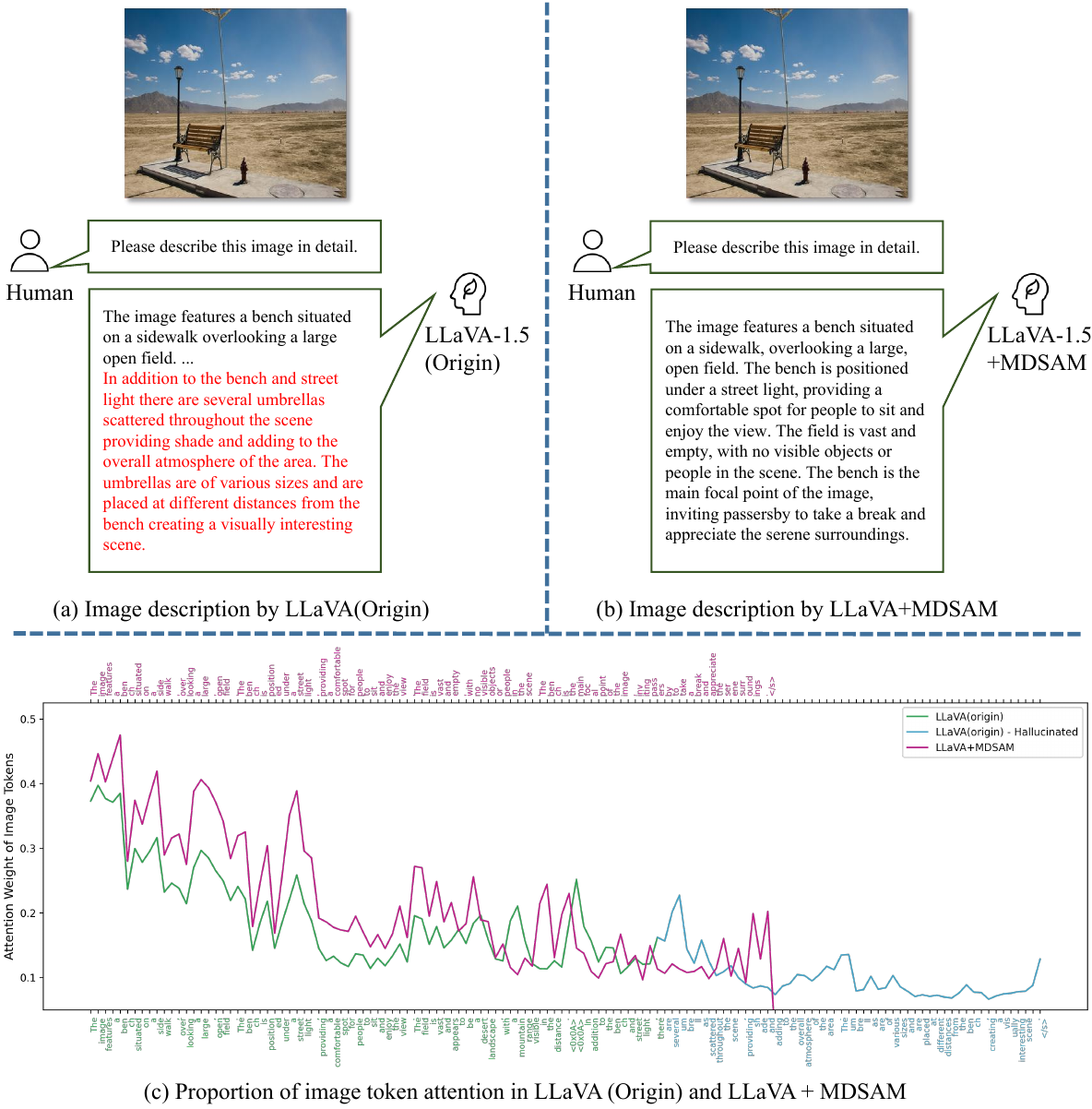}
  \caption{A example illustrating the comparison of image description and attention dynamics between LLaVA (Origin) and LLaVA+MDSAM.
  (a) The image description generated by LLaVA-1.5 includes hallucinated details, such as "several umbrellas scattered throughout the scene," which are not present in the actual image.
  (b) The description generated by LLaVA-1.5 + MDSAM is more accurate and concise, focusing on the bench, streetlight, and open field without introducing hallucinations.
  (c) The proportion of attention weights allocated to image tokens during generation is shown. Attention peaks when generating real or hallucinated objects. With LLaVA (Origin) , low attention on image tokens leads to hallucinations. In contrast, LLaVA+MDSAM increases visual attention, reducing hallucinations and focusing on salient visual information without introducing new noise. This demonstrates that MDSAM enhances the model's ability to prioritize relevant details while maintaining attention stability.}
  \label{fig1}
\end{figure}

Several researchers have proposed optimization strategies, such as Internvl(\cite{chen2024internvl}), LLaVA-RLHF(\cite{sun2023aligning}), and OPERA(\cite{huang2024opera}). However, these strategies all require additional training to mitigate hallucinations. We present a scenario where LVLMs generate hallucinated object descriptions. Specifically, during the decoding process, the generation of entities is highly sensitive to attention on the image, as illustrated in Fig \ref{fig1}.

As shown in Fig \ref{fig1}, when generating more tokens, the model becomes more sensitive to image tokens when producing real entities such as "bench" and "sidewalk," as well as hallucinated entities like "umbrella." This sensitivity is reflected in the peak points observed in Fig \ref{fig1}(c). To address this, we propose a novel training-free strategy called Memory-Driven Sparse Attention Matrix (MDSAM). MDSAM memorizes the attention patterns of image tokens at each layer and activates them through alignment during decoding. Specifically, it integrates with the corresponding layer weights at each layer to adjust the distribution of image tokens dynamically. This enhances the model's focus on relevant image tokens, effectively mitigating hallucinations.

To evaluate the effectiveness of MDSAM, we assessed response accuracy in image captioning tasks using CHAIR (\cite{radford2021learning}) and MME. Additionally, we employed POPE (\cite{li2023evaluating}) and MMHAL-BENCH (\cite{sun2023aligning}) to comprehensively evaluate the model's hallucination performance on VQA tasks. Since our method intervenes during the inference process, it can be applied to any decoding approach. Therefore, we conducted experiments across different models. The results demonstrate the efficacy of our method in mitigating hallucinations. Specifically, when applied to LLaVA-1.5-7B, MDSAM outperforms the original model on the CHAIR benchmark, reducing CHAIR$ _S $ by 10.6 and CHAIR$ _I $ by 3.2.

To summarize, the primary contributions of this work are as follows:

\begin{enumerate}
\item We introduce a novel training-free approach, MDSAM, which dynamically captures the attention LVLMs allocate to image tokens during decoding and activates updates to refine this attention at each layer. 

\item MDSAM is compatible with a wide range of LVLMs, such as LLaVA-1.5-7B, MiniGPT4 and DeepseekVL-7B, showcasing its adaptability across diverse LVLM architectures.  

\item We performed comprehensive experiments, including image captioning and visual question answering tasks, to assess the performance of MDSAM across various datasets and models. The findings validate the effectiveness of our method in reducing hallucinations.
\end{enumerate}

\section{Related Work}
\subsection{Large Vision-Language Models.}

LVLMs have made remarkable progress by leveraging advanced large language models(LMMs)(\cite{gilardi2023chatgpt,touvron2023llama,chiang2023vicuna,bai2023qwen,touvron2023llama2}) and seamlessly integrating information from multiple modalities, such as text, images, and audio, to understand and generate diverse content. Notable examples include Flamingo(\cite{alayrac2022flamingo}), LLaVA(\cite{liu2023visual,liu2024improved}), Qwen-VL(\cite{bai2023qwenvl}), Shikra(\cite{chen2023shikra}), InternVL(\cite{chen2024internvl}), MiniGemini(\cite{li2024minigemini}), MiniGPT4(\cite{zhu2023minigpt}), and DeepseekVL(\cite{lu2024deepseek}). However, similar to LLMs, LVLMs still face significant challenges related to the generation of hallucinations(\cite{liu2023mitigating,li2023evaluating}).

\subsection{Mitigating Hallucinations in LVLMs}
Although several existing studies have explored the issue of hallucinations in LVLMs and proposed various methods to mitigate their impact, different approaches have been developed. (\cite{chen2024internvl}) enhanced the connection between visual features and text by leveraging LLaMA2 to create QLLaMA. LLaVA-HACL (\cite{jiang2024hallucination}) observed a significant gap between visual and textual tokens in LVLMs, reducing hallucinations by explicitly adding new learning objectives to align these modalities. Similarly, LLaVA-RLHF (\cite{sun2023aligning}) adopted reinforcement learning strategies to better integrate visual and textual information. Direct Preference Optimization (DPO) (\cite{rafailov2023direct}) trains models directly from human preference data, avoiding RLHF(\cite{stiennon2020learning}) complexities. Using hallucination-labeled datasets, variants like HA-DPO (\cite{zhao2023beyond}) and FDPO (\cite{gunjal2024detecting}) improve alignment with human preferences, reducing hallucinatory outputs. OPERA(\cite{huang2024opera}) noted that LVLMs may hallucinate by over-focusing on a few abstract tokens while neglecting image tokens. To address this, OPERA modifies the beam search process with a weighted scoring system to deprioritize over-trusted candidates and introduces a retrospection mechanism to re-evaluate key decision points. 

In contrast, our method dynamically memorizes the sparse attention LVLMs pay to image tokens during decoding and activates the attention matrix for image tokens at each layer to update their distribution. Compared to other approaches, our method not only reduces computational costs but also achieves superior performance.

·\section{Preliminary}
\textbf{The architecture of LVLMs.}
LVLMs are typically composed of the following three components: a visual encoder, a connector (or projector) for modal alignment, and an advanced language model backbone such as an LLM. Both the visual encoder and the language model are usually pre-trained. During the pre-filling stage, the visual encoder often based on architectures like ViT. The input image $ I $ into a visual representation $ I_v $, which is then projected into the textual space by the connector, resulting in $ N_i $ image tokens $ X_i \in \mathbb{R}^{N_i \times d} $. Concurrently, the LLM tokenizer encodes the text prompt into $ N_p $ textual tokens $ X_t \in \mathbb{R}^{N_t \times d} $. These visual and textual tokens are concatenated to form the complete input sequence $ X = [X_i, X_t] \in \mathbb{R}^{(N_i + N_t) \times d} $, which is passed to the language decoder. The decoder then generates appropriate responses based on the provided instructions, enabling seamless integration of visual and textual information for multimodal tasks.

\textbf{Multi-head Attention in Transformers.}
Most LVLMs utilize models from the LLaMA family as their language decoders, which rely on the self-attention mechanism. The visual tokens generated by the projector are combined with text tokens and subsequently fed into the LLaMA model for forward decoding. From the perspective of a single attention head within a single layer, each head repeatedly performs an attention operation, maintaining the same input shape throughout:

\begin{equation}
O = A_hW_o, \quad A_h = \text{Attention}(Q, K) = \text{softmax}\left(\frac{Q_h K_h^\top}{\sqrt{d_k}}\right)
\end{equation}

Each attention head $ h $ performs an attention operation using its unique set of queries $ Q_h \in \mathbb{R}^{n \times d_k} $, keys $ K_h \in \mathbb{R}^{n \times d_k} $, and values $ W_o \in \mathbb{R}^{n \times d_k} $, where $ n $ denotes the sequence length and $ d_k $ represents the hidden dimensions. The output $ O \in \mathbb{R}^{n \times d_k} $ is computed by multiplying the values $ W_o $ with the attention weights $ A_h \in \mathbb{R}^{n \times n} $, where each row of $ A_h $ corresponds to the attention weights for each token during feature mixing. This mechanism allows the model to focus on different parts of the input sequence for each head, capturing diverse aspects of information from the token representations through the attention weights.
\begin{figure}
  \centering
  \includegraphics[width=\textwidth]{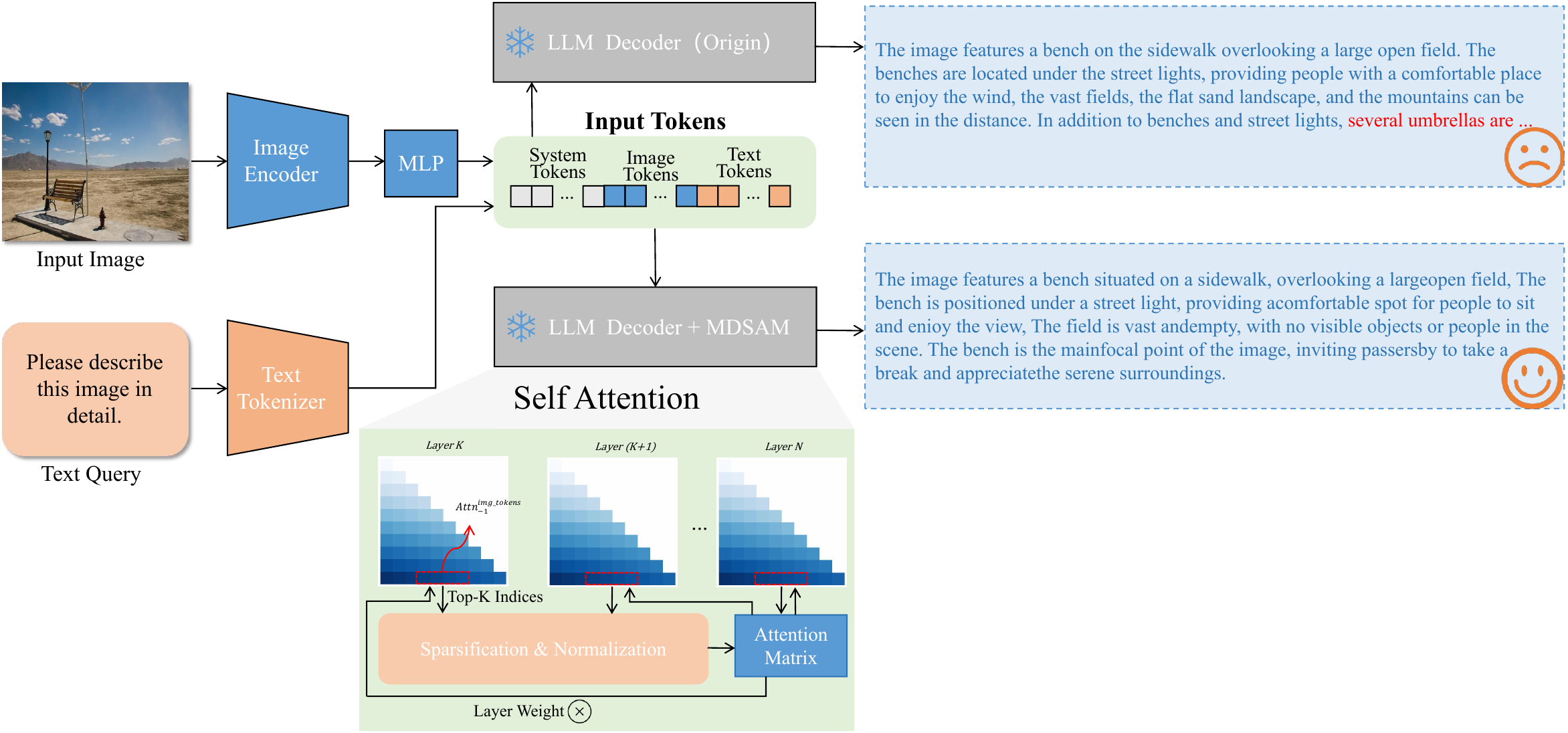}
  \caption{Architecture of our MDSAM}
  \label{fig2}
\end{figure}

\section{Method}
 Our approach iS focus on addressin the issues of image neglect and over-attention, which are fundamentally interconnected. Essentially, both increased attention to irrelevant parts of the image and neglect of important parts stem from noise present in each layer. By learning the attention matrix for image tokens from previous layers at each layer, we can reduce hallucinations in that layer. Additionally, to prevent excessive influence from earlier layers on the current layer, we apply sparsification to the matrix, ensuring it has a significant impact on the response. This strategy promotes more image-centric latent representations.
\subsection{Memory-Driven Sparse Attention Matrix}
We analyze the self-attention mechanism from a token-level perspective. In LVLMs, responses are generated token by token, conditioned on the input image, instruction, and previously generated tokens. This process relies on a multi-layer attention decoder, producing a probability distribution for the current token. Our goal is to extract the attention matrix of each head at every layer to capture the influence of different content during inference.

As can be seen from Fig \ref{fig1}, there is a certain correlation between the attention on images and the probability of entity generation. Therefore, we can enhance the probability of entity generation by strengthening the image attention at each layer. However, this approach may lead to an increase in noise, which prevents hallucinations from being effectively mitigated. To achieve more reliable attention, we construct a sparse attention matrix $ A $. At the $ l $-th layer, the image attention of the hidden state of the last token is first sparsified using a top-$ k $ selection process, and then updated into the attention weight matrix $ A $. Additionally, to reduce memory usage while maintaining high accuracy, we can choose an appropriate $ L $. The sparse attention matrix at the $ l $-th layer can be expressed as follows:

\begin{align}
   A_{\text{sparse}}^l = Contact\left(A_{\text{sparse}}^{l-1}[1:L], \, {{topk}({Normalize}}(\left| {A}_{\text{attn}}[n, i_{\text{start}}:i_{\text{end}}]\right|)) \right) 
\end{align}    
Here, $ A_{\text{sparse}}^{l-1} [1:] $ represents the retained portion of the sparse attention matrix from the previous layer, excluding the oldest entry. This memory-driven approach ensures that past attention patterns are preserved while limiting storage and computational costs. The function $\text{Normalize}(\cdot)$ is applied to rescale the attention weights of the last token $ n $ on the image tokens $[i_{\text{start}} : i_{\text{end}}]$, ensuring that they are well-behaved and interpretable. Specifically, normalization is performed as follows:
\begin{align}
\text{Normalize}(\mathbf{V}) = \frac{v - \min(\mathbf{V})}{\max(\mathbf{V}) - \min(\mathbf{V})},
\end{align}
where $ v $ represents individual elements of the attention weight matrix $ \mathbf{V} $. This min-max normalization scales the attention weights to the range $[0, 1]$, stabilizing the attention mechanism and preventing any single dimension from dominating.

After normalization, the $\text{topk}(\cdot)$ function performs a top-$k$ selection on the normalized attention weights corresponding to the image tokens in the current layer. This process retains only the most significant dimensions, effectively filtering out irrelevant or noisy features. The top-$k$ selection is defined as:

\begin{align}
{topk}(V, \tau) = V \cdot  \arg\min_{\mathbf{X}} \left\| \sum_{i=1}^m \sum_{j=1}^n X_{ij} - k \right\|\\ 
where\quad X_{ij} \in \{0, 1\},  \quad k = \tau \cdot (i_{end} - i_{start} + 1 )\nonumber
\end{align}
where $ X_{ij} \in \{0, 1\} $ and $ k = \tau \cdot (i_{\text{end}} - i_{\text{start}} + 1) $. This formulation ensures that only $ k $ elements are selected, determined by the threshold $ \tau $. The binary mask $ \mathbf{X} $ is optimized to minimize the difference between the sum of selected elements and $ k $, ensuring that the most important dimensions are retained. This approach effectively filters out irrelevant or noisy dimensions, improving the model's ability to focus on salient visual features.

·

\subsection{Attention Matrix Weight Activation and Regression}

To further enhance the model's ability to focus on salient visual features, we also introduce a memory-driven weighted aggregation mechanism. This mechanism leverages the sparse attention matrices $ A_{\text{sparse}} $ from multiple layers, combining them through a weighting scheme to capture both short-term and long-term attention patterns. The weights are designed to prioritize more recent layers while still retaining information from earlier layers, ensuring a balanced representation of attention dynamics across the network.The weighted aggregation of the sparse attention matrices is formalized as follows:
\begin{align}
\overline{A^l_{\text{mean}}} = \frac{\sum_{i=1}^{L} \left( A^l_{\text{sparse}} \cdot w_i \right)}{\sum_{i=1}^{L} w_i} = \frac{\sum_{i=1}^{L} \left( A^l_{\text{sparse}} \cdot \alpha^i \right)}{\sum_{i=1}^{L} \alpha^i}
\end{align}
The weighted aggregation mechanism computes the mean aggregated attention matrix $\overline{A^l_{\text{mean}}}$ as a weighted sum of the sparse attention matrices $A^l_{\text{sparse}}$ across $L$ layers. The weights $w_i = \alpha^i$ are determined by an exponential decay factor $\alpha \in (0, 1)$, which assigns higher importance to more recent layers while still incorporating contributions from earlier layers. This design ensures that the model dynamically balances short-term and long-term attention patterns. The numerator, $\sum_{i=1}^{L} \left( A^l_{\text{sparse}} \cdot \alpha^i \right)$, computes the weighted contribution of each layer's sparse attention matrix, emphasizing the most relevant and recent information. The denominator, $\sum_{i=1}^{L} \alpha^i$, normalizes the weights to ensure stability and explainability, preventing any single layer from disproportionately influencing the final representation. 


\subsection{Image Attention Alignment}
After completing Attention Matrix Weight Activation and Regression, we propose a method to further refine the attention weights assigned to image tokens in order to improve the alignment between visual and textual information. This is achieved by incorporating the aggregated attention matrix $\overline{A^l_{\text{mean}}}$ into the original attention weights. Specifically, the updated attention weights for the image tokens are computed as a weighted combination of the original attention weights and the aggregated attention matrix.The update rule for the attention weights is formalized as follows:
\begin{align}
    A^{L_{new}}_{\text{attn}}[n, i_{\text{start}} : i_{\text{end}}] = \frac{A_{\text{attn}}[n, i_{\text{start}} : i_{\text{end}}] + \beta \cdot \overline{A^l_{\text{mean}}}}{1 + \beta}
\end{align}

Here, $A_{\text{attn}}[n, i_{\text{start}} : i_{\text{end}}]$ represents the original attention weights for the image tokens, and $\overline{A^l_{\text{mean}}}$ is the aggregated attention matrix obtained through the memory-driven weighted aggregation mechanism. The hyperparameter $\beta$ controls the influence of the aggregated attention matrix on the final result, allowing for flexible adjustment based on the task requirements. 

This refinement enhances the model's ability to focus on salient visual features while maintaining stability and interpretability. By combining the original attention weights with the aggregated matrix, the model effectively increases the attention on entity-related image regions while avoiding excessive enhancement of non-entity regions. This targeted adjustment helps mitigate hallucinations, as it reduces the likelihood of the model generating descriptions for irrelevant or non-existent objects in the image.

\section{Experiments}



\subsection{Setup}
\label{subsec1}
\textbf{Baselines.}
We evaluate the effectiveness of our method on three different models to analyze the impact of image feature tokens processed by different projectors. To this end, we select two models that use linear projectors, LLaVA and DeepSeekVL, as well as one model that employs resamplers, MiniGPT4. 
Additionally, we include comparisons with OPERA(\cite{huang2024opera}), an advanced variant of beam search, and VCD(\cite{leng2024mitigating}), an improved version of nucleus sampling, to further validate our approach. We also incorporate AGLA(\cite{an2024agla}), which enhances the model’s visual perception by integrating both local and global image features. All experiments are conducted using the default hyperparameters from the open-source implementations of these methods.

\textbf{Implementation Details.}
We implement and evaluate the MDSAM method on three models: {LLaVA-1.5-7B}, {DeepSeekVL-7B}, and {MiniGPT4}. The size of the attention sparse matrix is configured as $ L = 8 $, ensuring that sufficient information from earlier layers is captured while maintaining manageable memory usage. For the hyperparameters $ \tau $, $ \alpha $, and $ \beta $, we assign the following values: for {LLaVA}, $[\tau=0.7, \alpha=0.9, \beta=0.6]$; for {DeepSeekVL}, $[\tau=0.8, \alpha=0.9, \beta=0.5]$; and for {MiniGPT4}, $[\tau=0.6, \alpha=0.9, \beta=0.5]$. In comparative experiments, we utilize the default parameter settings provided by the respective implementations of the baseline methods.


\subsection{Benchmarks and Metrics}
\textbf{CHAIR(\cite{rohrbach2018object}).}
The Caption Hallucination Assessment with Image Relevance  measures object hallucinations in image captions by comparing the objects mentioned in the generated text to the ground-truth objects present in the image. To evaluate the quality of the generated captions, we utilize three key metrics: CHAIRI (instance-level hallucination rate), CHAIRS (sentence-level hallucination rate), and Recall (coverage of ground-truth objects). These metrics collectively assess both the accuracy and relevance of the model's outputs.

\textbf{POPE(\cite{li2023evaluating}).} The Polling-based Object Probing Evaluation (POPE) is a dataset specifically created to assess object-level hallucinations in question-answering tasks. It features a collection of true/false questions related to images, such as "Is there a dog in the picture?". Using a dataset of images and their associated object annotations, POPE generates triples consisting of an image, a corresponding question, and an answer. This structured approach enables the evaluation of how accurately models respond to queries about objects present in the images.

\textbf{MME evaluation\cite{fu2024mmecomprehensiveevaluationbenchmark}}
MME serves as a comprehensive framework for assessing multimodal large language models, focusing on both perceptual and cognitive abilities. It includes 14 subtasks, such as object existence, counting, position, color recognition, OCR, commonsense reasoning, numerical calculation, text translation, and code reasoning. These tasks are grouped into four categories: Reasoning Tasks (commonsense reasoning, numerical calculation, text translation, code reasoning), Perception Tasks (existence, count, position, color), OCR Tasks (optical character recognition), and Fine-Grained Tasks (poster, celebrity, scene, landmark, artwork interpretation). This structured categorization ensures a thorough evaluation of the models’ capabilities across diverse domains.

\textbf{MMHal-Bench(\cite{sun2023aligning})}
To further assess our method on more complex datasets, we utilize MMHal-Bench, which comprises 96 image-question pairs spanning 8 question categories and 12 object topics. The dataset features eight distinct types of questions, including those addressing object attributes, adversarial objects, comparisons, counting, spatial relationships, environmental context, holistic descriptions, and other challenging aspects. This setup is designed to thoroughly evaluate the model's hallucination performance on high-complexity tasks. Unlike POPE, which primarily focuses on existence-based questions, MMHal-Bench incorporates logical reasoning into its queries, making it a more nuanced VQA-based evaluation. For the assessment on MMHal-Bench, we first generate answers to the questions and then employ GPT-4 to assign scores by comparing the responses with the ground-truth answers. The evaluation results provide scores for each question category, with the overall score calculated as the average of these individual scores.

\subsection{Experimental Results}


\begin{table}
\centering
\caption{\textbf{Evaluation results of CHAIR on three LVLMs.} CHAIR is used as the metric, where lower scores indicate reduced hallucinations. Additionally, we report the average Recall and the average caption length to provide a comprehensive comparison across models.}
\begin{tabular}{llcccc} 
\toprule
\textbf{Model}              & \textbf{Method~ ~~} & \multicolumn{1}{l}{CHAIRS↓} & \multicolumn{1}{l}{CHAIRI↓} & \multicolumn{1}{l}{Recall↑} & \multicolumn{1}{l}{Avg. Len}  \\ 
\hline
\multirow{2}{*}{LLaVA-1.5}  & Greedy              & 48.4                        & 13.0                        & 76.9                        & \textbf{96.9}                 \\
                            & MDSAM               & \textbf{37.8}               & \textbf{9.8}                & \textbf{77.0}               & 94.2                          \\ 
\hline
\multirow{2}{*}{MiniGPT4}   & Greedy              & 35.6                        & 9.9                         & \textbf{61.7}               & 89.4                          \\
                            & MDSAM               & \textbf{26.0}               & \textbf{9.1}                & 57.9                        & \textbf{98.8}                 \\ 
\hline
\multirow{2}{*}{DeepseekVL} & Greedy              & 25.0                        & 6.1                         & \textbf{63.1}               & 105.7                         \\
                            & MDSAM               & \textbf{21.8}               & \textbf{5.2}                & 59.8                        & \textbf{107.3}                \\
\hline
\end{tabular}
\label{table1}
\end{table}

\textbf{CHAIR evaluation on hallucinations.}
we evaluate the performance of multimodal large language models (LVLMs) using the CHAIR metric, which measures hallucinations in image captioning tasks by comparing generated captions to ground-truth object annotations. Lower scores for CHAIRS (sentence-level hallucination rate) and CHAIRI (instance-level hallucination rate) indicate fewer hallucinations, while higher Recall scores reflect better alignment with ground-truth objects. We also report the average caption length (Avg. Len ) to analyze the trade-off between detail and hallucination reduction.

\begin{table}
\centering
\caption{\textbf{Comparison of CHAIR performance between MDSAM and other approaches on LLaVA-1.5.}}
\begin{tabular}{lccc} 
\hline
Method & \multicolumn{1}{l}{CHAIRS} & \multicolumn{1}{l}{CHAIRI} & \multicolumn{1}{l}{Recall}  \\ 
\hline
Greedy & 48.4                       & 13.0                       & 76.9                        \\
Beam   & 53.1                       & 13.6                       & \textbf{77.6}               \\
OPERA  & 45.5                       & 14.6                       & 76.5                        \\
VCD    & 52.7                       & 13.8                       & 77.5                        \\
AGLA   & 49.8                       & 13.9                       & 77.1                        \\
MDSAM  & \textbf{37.8}              & \textbf{9.8}               & 77.0                        \\
\hline
\end{tabular}
\label{table2}
\end{table}
Our experiments involve three LVLMs: LLaVA-1.5, MiniGPT4, and DeepseekVL, evaluated under two decoding methods: Greedy and MDSAM . In Table \ref{table1} Across all models, MDSAM consistently reduces hallucinations, achieving significant improvements in both CHAIRS and CHAIRI scores. For instance, in LLaVA-1.5 , MDSAM reduces CHAIRS from 48.4 to 37.8 and CHAIRI from 13.0 to 9.8 while maintaining a competitive Recall score of 77.0. Similarly in Table \ref{table2}, MDSAM outperforms other decoding methods like Beam , OPERA , and VCD on LLaVA-1.5 , achieving the lowest hallucination rates without compromising overall caption quality. These results highlight the effectiveness of MDSAM in mitigating hallucinations while maintaining alignment with ground-truth information.

\begin{table}
\centering
\caption{\textbf{Evaluation results of POPE on three LVLMs. }}
\begin{tabular}{{lllllll}} 
\hline
Model  & \multicolumn{2}{l}{LLaVA-1.5}   & \multicolumn{2}{l}{MiniGPT4}    & \multicolumn{2}{l}{DeepseekVL}   \\ 
\hline

Method & Acc↑            & F1↑             & Acc↑            & F1↑             & Acc↑            & F1↑              \\ 
Greedy & 84.82          & 85.41          & \textbf{75.51} & 74.92          & 87.12          & 88.26           \\
MDSAM  & \textbf{85.24} & \textbf{85.56} & 75.31          & \textbf{75.03} & \textbf{87.61} & \textbf{88.45}  \\

\hline
\end{tabular}
\label{table3}
\end{table}
\textbf{POPE evaluation on hallucinations.}
In Table \ref{table3}, we compare the performance of various decoding-based hallucination mitigation methods on the POPE benchmark, where “Greedy” represents the baseline decoding strategy. Clearly, MDSAM outperforms the Greedy method across all LVLMs, achieving consistent improvements in both accuracy and F1 scores. For instance, in LLaVA-1.5, MDSAM increases accuracy from 84.82 to 85.24 and the F1 score from 85.41 to 85.56. Similarly, in DeepseekVL, MDSAM achieves notable gains, with accuracy rising from 77.74 to 78.61 and the F1 score improving from 78.32 to 78.95. Although the improvement in MiniGPT4 is marginal, with a slight decrease in accuracy (75.51 to 75.31) but a small increase in the F1 score (74.92 to 75.03), the overall results demonstrate the effectiveness of MDSAM in mitigating object hallucinations while maintaining or enhancing model performance. These findings highlight the robustness of MDSAM as a superior decoding strategy for improving the reliability of LVLMs.

\textbf{MME evaluation on hallucinations.}
In Table \ref{table4}, we present the MME evaluation results for three multimodal large language models—LLaVA-1.5, MiniGPT4, and DeepseekVL—on tasks categorized into Reasoning, OCR, Fine-Grained, and Coarse-Grained subtasks. The evaluation spans both cognitive and perceptual abilities, with scores aggregated across these domains. Notably, MDSAM consistently outperforms the Greedy decoding method across all models, achieving higher total scores and demonstrating its effectiveness in mitigating hallucinations while enhancing overall performance. For instance, in LLaVA-1.5 , MDSAM improves the total score from 1505.18 to 1518.88, with consistent gains in both cognitive (345.36 to 355.71) and perceptual (137.5 to 147.5) subtasks. Similarly, in DeepseekVL , MDSAM achieves a marginal but consistent improvement, increasing the total score from 1551.30 to 1555.40. These results highlight that MDSAM not only enhances reasoning and fine-grained perception but also maintains robustness across coarse-grained tasks, underscoring its ability to reduce hallucinations and improve the reliability of multimodal models.

\begin{table}
\caption{\textbf{Evaluation results of MME on three LVLMs.}}
\begin{adjustbox}{width=\textwidth,center} 
\begin{tabular}{llcccccc} 
\hline
\multirow{2}{*}{Model}      & \multirow{2}{*}{Method} & Cognition score & \multicolumn{4}{c}{Perception Scorel}                                                                                                   \\ 
\cmidrule(r){3-3}\cmidrule(r){4-7}
                            &                         & Reasoning Tasks & OCR Task        & Fine-Grained Tasks & Coarse-Grained Tasks                                                 & Total                     \\ 
\hline
\multirow{2}{*}{LLaVA-1.5}  & Greedy                  & 345.36          & 137.5           & \textbf{724.35}    & 643.33                                                               & 1505.18                   \\
                            & MDSAM                   & \textbf{355.71} & \textbf{147.5}  & 723.05             & \textbf{648.33}                                                      & \textbf{1518.88}          \\
\multirow{2}{*}{MiniGPT4}   & Greedy                  & 144.29          & 57.5            & 282.49             & {\textbf{241.67}} & 581.66                    \\
                            & MDSAM                   & \textbf{145.99} & \textbf{69.5}   & \textbf{283.54}    & 240.61                                                               & {\textbf{593.65}}  \\
\multirow{2}{*}{DeepseekVL} & Greedy                  &363.70          &150.16          & 738.71             & 662.43                                                               & 1551.30                   \\
                            & MDSAM                   & \textbf{365.22} & \textbf{151.30} & \textbf{740.15}    & \textbf{663.95}                                                      & \textbf{1555.40}          \\
\hline
\end{tabular}
\end{adjustbox}
\label{table4}
\end{table}

\begin{table}
\centering
\caption{\textbf{Evaluation results of MMHal-Bench on three LVLMs.} }
\begin{adjustbox}{width=\textwidth,center} 
\begin{tabular}{llcccccccccc} 
\hline
\multirow{2}{*}{Model}      & \multirow{2}{*}{Method} & \multicolumn{1}{l}{\multirow{2}{*}{Average score↑ }} & \multicolumn{1}{l}{\multirow{2}{*}{Hallucination rate↓ }} & \multicolumn{8}{c}{Score in Each Question Type ↑}                                                                                                                                                                                                            \\ 
\cmidrule(lr){5-12}
                            &                         & \multicolumn{1}{l}{}                                 & \multicolumn{1}{l}{}                                      & \multicolumn{1}{l}{Attribute} & \multicolumn{1}{l}{Adversarial} & \multicolumn{1}{l}{Comparison} & \multicolumn{1}{l}{Counting} & \multicolumn{1}{l}{Relation} & \multicolumn{1}{l}{Environment} & \multicolumn{1}{l}{Holistic} & \multicolumn{1}{l}{Other}  \\ 
\hline
\multirow{2}{*}{LLaVA-1.5}  & Greedy                  & 2.18                                                 & 0.64                                                      & 2.33                          & \textbf{1.92}                   & 2.83                           & 2.17                         & \textbf{2.17}                & \textbf{3.42}                   & 1.33                         & 1.25                       \\
                            & MDSAM                   & \textbf{2.41}                                        & \textbf{0.58}                                             & \textbf{3.17}                 & 1.83                            & \textbf{3.75}                  & \textbf{2.33}                & 2.0                          & 3.08                            & \textbf{1.83}                & 1.25                       \\
\multirow{2}{*}{MiniGPT4}   & Greedy                  & 1.40                                                 & 0.76                                                      & 0.79                          & 1.85                            & 2.18                           & 0.93                         & 1.25                         & 1.36                            & 0.93                         & \textbf{1.92}              \\
                            & MDSAM                   & \textbf{2.04}                                        & \textbf{0.69}                                             & \textbf{1.33}                 & \textbf{1.98}                   & \textbf{2.35}                  & \textbf{1.86}                & \textbf{2.66}                & \textbf{2.98}                   & \textbf{1.35}                & 1.77                       \\
\multirow{2}{*}{DeepseekVL} & Greedy                  & 2.15                                                 & 0.62                                                      & 2.45                          & 1.87                            & 2.92                           & 1.95                         & 1.85                         & \textbf{3.20}                   & 1.32                         & \textbf{1.66}              \\
                            & MDSAM                   & \textbf{2.39}                                        & \textbf{0.55}                                             & \textbf{2.98}                 & \textbf{1.89}                   & \textbf{3.15}                  & \textbf{2.12}                & \textbf{2.65}                & 3.01                            & \textbf{1.64}                & 1.64                       \\
\hline
\end{tabular}
\end{adjustbox}
\label{table5}
\end{table}

\textbf{MMHal-Bench evaluation on hallucinations.} In Table \ref{table5} , we present the evaluation results of MMHal-Bench on three state-of-the-art Large Vision-Language Models (LVLMs), focusing on their performance in mitigating hallucinations. The table provides a detailed breakdown of average scores, hallucination rates, and performance across eight distinct question types. Notably, the MDSAM method consistently outperforms the Greedy approach across all models, achieving higher average scores and lower hallucination rates. For instance, LLaVA-1.5 with MDSAM achieves an average score of 2.41 and a hallucination rate of 0.58, surpassing its Greedy counterpart (2.18 and 0.64, respectively). Similar improvements are observed for MiniGPT4 and DeepseekVL, highlighting the effectiveness of MDSAM in reducing erroneous outputs while enhancing overall accuracy. Among the question types, adversarial and holistic questions remain particularly challenging, as indicated by their relatively lower scores. However, the consistent gains achieved by MDSAM in these categories demonstrate its robustness and potential to address complex hallucination scenarios. 

\begin{table}
\centering
\caption{\textbf{Ablation Study of the Hyperparameter $\beta$,$\tau$}}
\begin{tabular}{cccccc} 
\hline
\multicolumn{1}{l}{\multirow{2}{*}{$\beta$}} & \multicolumn{1}{l}{\multirow{2}{*}{$\tau$}} & \multicolumn{4}{c}{\textbf{LLaVA-1.5}}  \\ 
\cmidrule(lr){3-6}
\multicolumn{1}{l}{}                   & \multicolumn{1}{l}{}                   & CHAIRS & CHAIRI & Recall & Avg. Len      \\ 
\hline
-                                      & -                                      & 48.4   & 13.0   & 76.2   & 96.9          \\ 
\hline
0.5                                    & 0.2                                    & 32.2   & 8.3    & 68.7   & 88.1          \\
0.5                                    & 0.4                                    & 37.8   & 9.8    & 72.8   & 94.2          \\
0.5                                    & 0.6                                    & 38.8   & 9.8    & 73.1   & 93.1          \\
0.5                                    & 0.8                                    & 41.8   & 10.6   & 74.0   & 95.9          \\
0.5                                    & 1.0                                    & 43.6   & 11.1   & 75.7   & 94.7          \\
1.0                                    & 1.0                                    & 32.8   & 8.4    & 69.4   & 89.2          \\
1.5                                    & 1.0                                    & 37.4   & 10.3   & 64.8   & 88.1          \\
2.0                                    & 1.0                                    & 44.0   & 14.9   & 61.5   & 88.6~         \\
\hline
\end{tabular}
\label{table6}
\end{table}

\subsection{Ablation Study}

Our method, MDSAM , consists of two intervention stages designed to mitigate hallucinations and enhance the quality of long-sequence image descriptions. In the first stage, during forward inference, we employ the hyperparameter $\tau$ to define the range for attention selection, ensuring that only the most salient dimensions in the attention matrix are retained through a structured sparsification process. In the second stage, we refine the attention weights by proportionally combining the original attention weights with the updated ones, a process governed by the parameter $\beta$ . This two-stage mechanism ensures both noise reduction and enhanced alignment between textual outputs and visual content.

To systematically evaluate the impact of hyperparameters on long-sequence image description tasks, we adopt LLaVA-1.5 as the representative LVLM baseline and use the greedy decoding method as the foundational comparison framework. For performance assessment, we utilize the CHAIR metric, which quantifies hallucination rates in generated captions. However, since CHAIR primarily focuses on measuring hallucinations, we complement it with the Recall score to account for informativeness and accuracy. The Recall score is computed by comparing the objects mentioned in the generated description against the ground-truth object set, while also considering hallucinated objects. This dual evaluation framework allows for a fair comparison: when Recall scores are similar across methods, the CHAIR metric provides a reliable indicator of hallucination mitigation effectiveness.

\textbf{Ablation Study of the Hyperparameter $\beta$, $\tau$.} 
To examine the influence of the hyperparameters $\beta$ and $\tau$ on model performance, we conduct an ablation study using LLaVA-1.5 as the baseline. The results in Table \ref{table6} demonstrate that $\tau$, which governs attention sparsity during the first stage, plays a critical role in balancing hallucination mitigation and informativeness. Smaller values of $\tau$ effectively reduce hallucinations, but can lead to less detailed descriptions, while larger values improve recall at the expense of increased hallucination rates. Similarly, $\beta$, which controls the blending ratio of original and updated attention weights in the second stage, exhibits a trade-off between refinement precision and robustness. Lower $\beta$ values tend to suppress hallucinations more effectively, whereas higher values prioritize richer but noisier outputs. These findings underscore the importance of tuning $\beta$ and $\tau$ to achieve an optimal balance between reducing hallucinations and maintaining descriptive quality.

\section{Conclusion}
In this paper, we first analyze the causes and manifestations of hallucinations in LVLMs, identifying that the generation of hallucinated entities is closely related to the model's sensitivity to image attention. This issue fundamentally arises from the model's dependency on image tokens during decoding. To address this, we introduce MDSAM, a novel training-free approach that dynamically captures the attention LVLMs allocate to image tokens during decoding and activates updates to refine this attention at each layer. MDSAM intervenes in the model's inference process, steering it toward a more image-grounded and trustworthy direction without requiring any external tools or additional training. Extensive experiments across multiple benchmarks and LVLMs demonstrate the effectiveness of MDSAM in mitigating hallucinations, showcasing its potential as a robust solution for enhancing the reliability and accuracy of multimodal language models.
\bibliographystyle{elsarticle-harv} 
\bibliography{cas-refs}






\end{document}